\documentclass[10pt,twocolumn,letterpaper]{article}

\usepackage{cvpr}      
\definecolor{cvprblue}{rgb}{0.21,0.49,0.74}
\usepackage[pagebackref,breaklinks,colorlinks,allcolors=cvprblue]{hyperref}

\usepackage{algorithm}
\usepackage{algorithmic}
\usepackage{amsmath}
\usepackage{bm}
\usepackage{amssymb}
\usepackage{cuted}
\usepackage{newfloat}
\usepackage{listings}
\usepackage[capitalize,noabbrev]{cleveref}
\usepackage{booktabs}       
\usepackage{multirow}       
\usepackage{color}
\usepackage[table]{xcolor} 
\usepackage{placeins}
\usepackage[accsupp]{axessibility}

\def\paperID{981} 
\def\confName{CVPR}
\def\confYear{2026}

\title{UAVLight: A Benchmark for Illumination-Robust 3D Reconstruction \\ in Unmanned Aerial Vehicle (UAV) Scenes}

\author{
Kang Du$^{1\thanks{Equal contribution.}}$ \quad
Xue Liao$^{1,2\footnotemark[1]}$ \quad
Junpeng Xia$^{2}$ \quad
Chaozheng Guo$^{2}$ \quad
Yi Gu$^{1}$ \quad
Yirui Guan$^{3}$ \quad \\
Sheng Huang$^{2}$ \quad
Zeyu Wang$^{1,4\thanks{Corresponding author.}}$\\ \\
$^{1}$The Hong Kong University of Science and Technology (Guangzhou)\\
$^{2}$Meituan UAV\\
$^{3}$Beijing University of Chemical Technology\\
$^{4}$The Hong Kong University of Science and Technology\\
{\tt\small Project page: \url{https://uavlight.github.io/}}
}

\begin{document}
\maketitle

\begin{strip}
\begin{minipage}{\textwidth}\centering
\vspace{-1.2cm}
\includegraphics[width=0.98\textwidth]{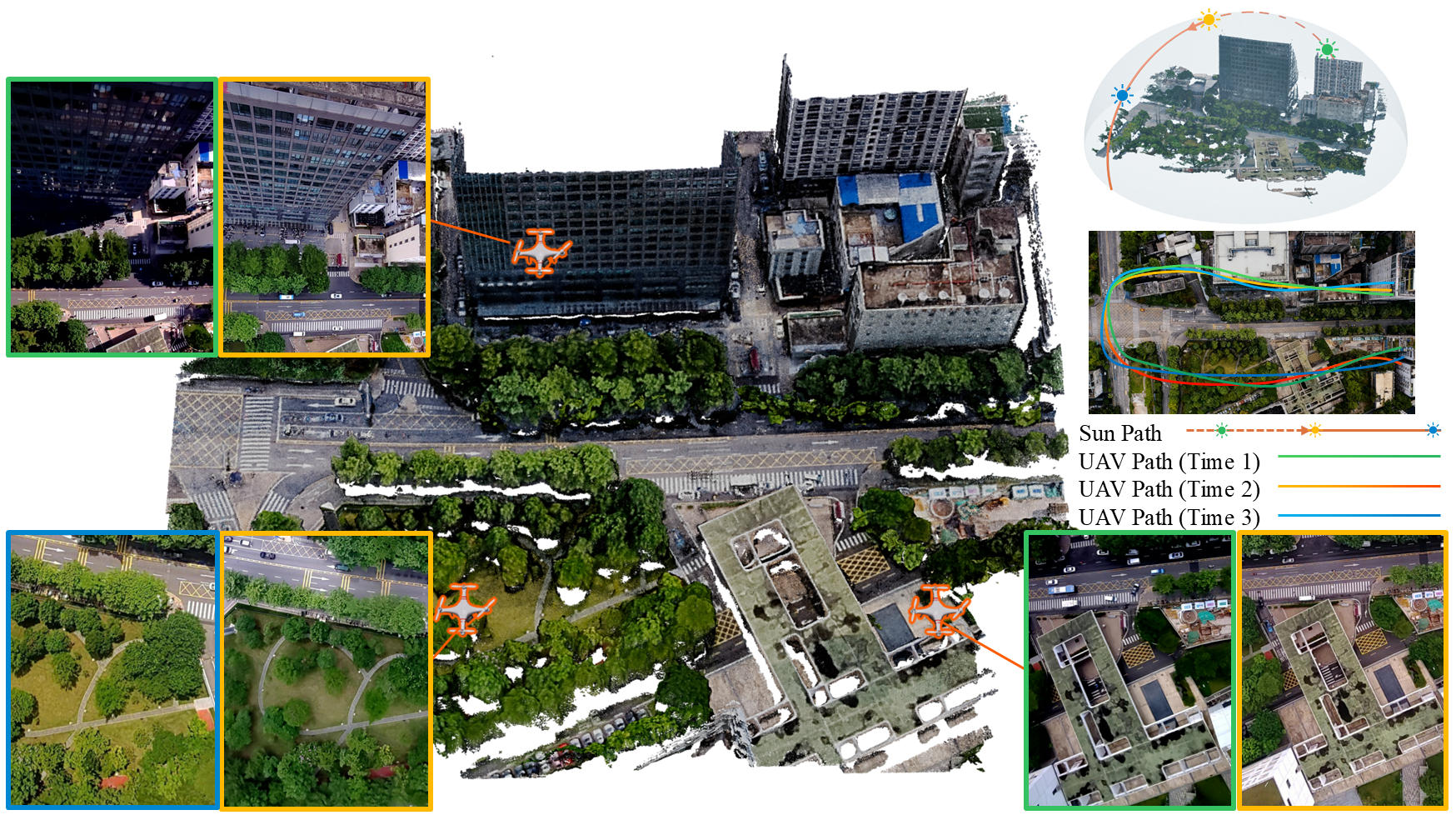}
\captionof{figure}{
Overview of the UAVLight benchmark.
Each scene is captured by low-altitude UAV flights along fixed waypointed trajectories at multiple times of day. 
Our benchmark records natural illumination changes along consistent geometry and viewpoints, enabling quantitative evaluation of illumination-robust reconstruction and relighting.
}
\label{fig:teaser}
\end{minipage}
\end{strip}

\begin{abstract}
Illumination inconsistency is a fundamental challenge in multi-view 3D reconstruction. Variations in sunlight direction, cloud cover, and shadows break the constant-lighting assumption underlying both classical multi-view stereo (MVS) and structure from motion (SfM) pipelines and recent neural rendering methods, leading to geometry drift, color inconsistency, and shadow imprinting. This issue is especially critical in UAV-based reconstruction, where long flight durations and outdoor environments make lighting changes unavoidable.
However, existing datasets either restrict capture to short time windows, thus lacking meaningful illumination diversity, or span months and seasons, where geometric and semantic changes confound the isolated study of lighting robustness.
We introduce UAVLight, a controlled-yet-real benchmark for illumination-robust 3D reconstruction. Each scene is captured along repeatable, geo-referenced flight paths at multiple fixed times of day, producing natural lighting variation under consistent geometry, calibration, and viewpoints. With standardized evaluation protocols across lighting conditions, UAVLight provides a reliable foundation for developing and benchmarking reconstruction methods that are consistent, faithful, and relightable in real outdoor environments.
\end{abstract}
    
\section{Introduction}
\label{sec:intro}

Recent neural 3D reconstruction methods have advanced from classical SfM~\cite{schoenberger2016sfm} and MVS~\cite{Yao2020BlendedMVS, ReviewUAV3DRecon} to neural fields~\cite{Mildenhall2020NeRF, BenchmarkNeRF} and Gaussian Splatting~\cite{Kerbl2023GaussianSplatting, UAVReconGS}, achieving photorealistic rendering and accurate geometry from casual multi-view imagery. However, most widely used benchmarks~\cite{Knapitsch2017TanksTemples, Yao2020BlendedMVS, Barron2022MipNeRF360} implicitly assume stationary illumination: scenes are captured within minutes under nearly fixed lighting. In contrast, 3D reconstruction with UAV often lasts for hours or is conducted at various times of a day, during which the solar position, intensity, and atmospheric conditions vary significantly. Such non-stationary outdoor illumination deviates from the constant-lighting assumption, leading to geometry drift, view-dependent color shifts, shadow imprinting in reflectance, and unstable relighting.

Existing approaches can be broadly grouped into the following two categories. (1) \emph{Implicit appearance modeling} augments neural fields with per-view/per-ray latents to absorb exposure, white balance, shadows, and weather-induced variations~\cite{Sun2022NeILF, MartinBrualla2021NeRFW, Kulhanek2024WildGaussians, Zhang2024GSW}. While this improves robustness ``in the wild'', it offers limited physical interpretability and unreliable relighting. (2) \emph{Explicit lighting estimation} factorizes appearance into reflectance and illumination via inverse rendering, enabling physically grounded relighting and shadow reasoning, but it requires strong priors (e.g., sun–sky models~\cite{Gardner2024NeuSky}), accurate calibration, and is brittle under auto-exposure~\cite{Chen2022NeRFOSR, Li2024ReCap, lumigauss}.

Recent studies have integrated 3D reconstruction into UAV scenarios \cite{Smith2018AerialPathPlanning, Yan2021SamplingBasedUAV, Shang2023TopologyUAVMVS}. However, there remains no dataset that systematically elicits, isolates, and evaluates performance under time-varying sunlight while keeping geometry and viewpoints consistent. A few recent datasets capture multiple lighting conditions, but are either object-centric with limited geometric complexity \cite{Ummenhofer2024OWL} or span long temporal windows \cite{Chen2022NeRFOSR}, during which geometry, vegetation, and transient objects evolve alongside illumination. These uncontrolled variations confound illumination effects, hindering isolation, quantification, and fair comparison of illumination robustness across methods. Consequently, research progress on \emph{illumination-robust reconstruction} remains difficult to measure, and evaluations between implicit and explicit methods are often inconclusive.

In this work, we aim to bridge this gap by designing a controlled-yet-real benchmark that isolates illumination variation while preserving real-world UAV acquisition complexity.
To isolate illumination effects while preserving real-world realism, we adopt three principles:
(i) focus on outdoor UAV ground scenes primarily lit by sunlight, avoiding indoor or multi-source lighting confounders;
(ii) capture at consistent time slots over short periods (e.g., consecutive days) to reduce non-illumination changes such as layout, vegetation, and human activity;
(iii) repeat consistent, waypointed flight trajectories to obtain comparable viewpoints and scene coverage across captures.
Furthermore, nadir-oriented flight plans greatly limit sky pixels, reducing HDR sky ambiguity and improving cross-method comparability.

Guided by these principles, we introduce \emph{UAVLight}, a benchmark and dataset that supports 3D reconstruction under realistic, time-varying natural lighting while controlling geometry and viewpoint comparability. Importantly, UAVLight explicitly disentangles illumination variation from other real-world factors through controlled acquisition, enabling systematic evaluation of illumination robustness.
For reliable quantitative evaluation, we provide high-quality, geo-referenced point clouds as a geometric reference. These references are anchored by RTK-based camera priors and ground control points, and further validated via checkpoint-based survey measurements, achieving an average geometric error of approximately 10~cm across scenes, consistent with standard UAV photogrammetry accuracy.
Beyond raw data, UAVLight standardizes splits, tasks, and metrics to jointly assess geometry, cross-time photometric consistency, and relighting stability. In particular, we emphasize image-space and cross-illumination evaluation protocols, which are better suited for assessing appearance consistency under varying natural lighting, while still being grounded by metric-scale geometric references.
UAVLight advances illumination-robust 3D reconstruction by providing:
\begin{itemize}
    \item \textbf{A Controlled Yet Real-World Benchmark:} We establish a multi-time-of-day UAV benchmark in which each scene is captured under multiple illumination conditions along repeated GPS-guided trajectories. This design isolates time-varying outdoor illumination while preserving the realism and complexity of real-world scenes.
    
    \item \textbf{Reliable Geometric Reference:} We provide high-quality geo-referenced point clouds, enabling absolute and scene-scale evaluation of reconstruction accuracy across diverse outdoor environments.
    
    \item \textbf{Standardized Evaluation Protocols:} We introduce a unified evaluation protocol for jointly assessing reconstruction fidelity and cross-time consistency, and benchmark representative methods spanning implicit appearance modeling and explicit inverse-rendering paradigms.
\end{itemize}

\begin{table*}[t]
\centering
\caption{A taxonomy of existing datasets.}
\label{tab:dataset_taxonomy}
\small
\begin{tabular}{@{}lcccccc@{}}
\toprule
\textbf{Dataset} & \textbf{Content} & \textbf{Purpose} & \textbf{Same-light} & \textbf{Light} & \textbf{\# Illum.} & \textbf{\# Scenes} \\
 &  &  & \textbf{(per seq.)} & \textbf{source} &  &  \\
\midrule
NeRF Synthetic\cite{Mildenhall2020NeRF} & Objects & Multi-view & - & Synthetic & - & 8 \\
TensoIR\cite{tensoir} & Objects & Multi-view & - & Synthetic & - & 4 \\
Objects with Lights\cite{Ummenhofer2024OWL} & Objects & Multi-view & Yes & Natural & 3 & 8 \\
OpenIllumination\cite{Oechsle2024OpenIllumination} & Objects & Multi-view & - & Light stage & 13 pattern+142OLAT & 64 \\
LSMI\cite{kim2021large} & Indoor & Single-view & - & Spotlight & 1-3 & 2700 \\
Multi-illumination in the Wild\cite{murmann2019dataset} & Indoor & Single-view & - & Spotlight & 25 & 1000 \\
Phototourism\cite{phototourism} & Outdoor & Multi-view & No & Natural & - & 13 \\
NeRF-OSR\cite{Chen2022NeRFOSR} & Outdoor & Multi-view & Yes & Natural & 5+ & 9 \\
\textbf{Ours} & \textbf{UAV} & \textbf{Multi-view} & \textbf{Yes} & \textbf{Natural} & \textbf{3-11} & \textbf{18} \\
\bottomrule
\end{tabular}
\end{table*}

\section{Existing Benchmarks}
\label{sec:existing_benchmarks}

As summarized in \Cref{tab:dataset_taxonomy}, existing illumination-aware 3D reconstruction datasets can be characterized along six axes of realism, controllability, and diversity: 
(1) content (indoor, object-centric, outdoor), 
(2) task (single- vs. multi-view), 
(3) lighting consistency within a sequence, 
(4) illumination source (synthetic, natural, controlled), 
(5) number of lighting conditions, and 
(6) number of scenes. 
Together, these axes form a concise taxonomy of current benchmarks.

Synthetic and object-centric datasets emphasize controlled appearance analysis. 
NeRF Synthetic~\cite{Mildenhall2020NeRF} and TensoIR~\cite{tensoir} provide clean geometry and known lighting, while LightCity~\cite{wang2025lightcity} extends this to urban scenes with multi-illumination; however, all lack real-world variability.
Object-centric datasets such as OWL~\cite{Ummenhofer2024OWL} and OpenIllumination~\cite{Oechsle2024OpenIllumination} support relighting and material analysis; OWL captures natural illumination with limited variation, while OpenIllumination uses light-stage setups with strong control but limited geometric complexity.

Indoor single-view datasets, e.g., LSMI~\cite{kim2021large} and Multi-Illumination Images in the Wild~\cite{murmann2019dataset}, focus on per-image illumination estimation under diverse artificial lighting. 
They lack multi-view consistency and stable geometry, limiting their use for reconstruction or cross-view evaluation.

Outdoor reconstruction datasets prioritize geometry but assume near-constant illumination. 
MipNeRF-360~\cite{Barron2022MipNeRF360} and Tanks\&Temples~\cite{Knapitsch2017TanksTemples} are captured within short time windows, resulting in limited lighting variation. 
Phototourism~\cite{phototourism} aggregates Internet photos with diverse viewpoints but uncontrolled lighting. 
NeRF-OSR~\cite{Chen2022NeRFOSR} introduces multiple lighting conditions, but long capture spans (months to years) introduce geometry and semantic changes, confounding illumination effects.

\textbf{Our Position.}
We propose an outdoor, scene-level, multi-view UAV dataset captured under natural sunlight, with consistent illumination within each sequence. 
Each scene includes 3--11 time slots collected along fixed waypointed trajectories across 18 scenes. 
This design preserves real-world realism while controlling confounding factors, enabling systematic evaluation of illumination robustness and relighting stability.

\begin{figure*}[!t]
\centering
\includegraphics[width=0.98\textwidth]{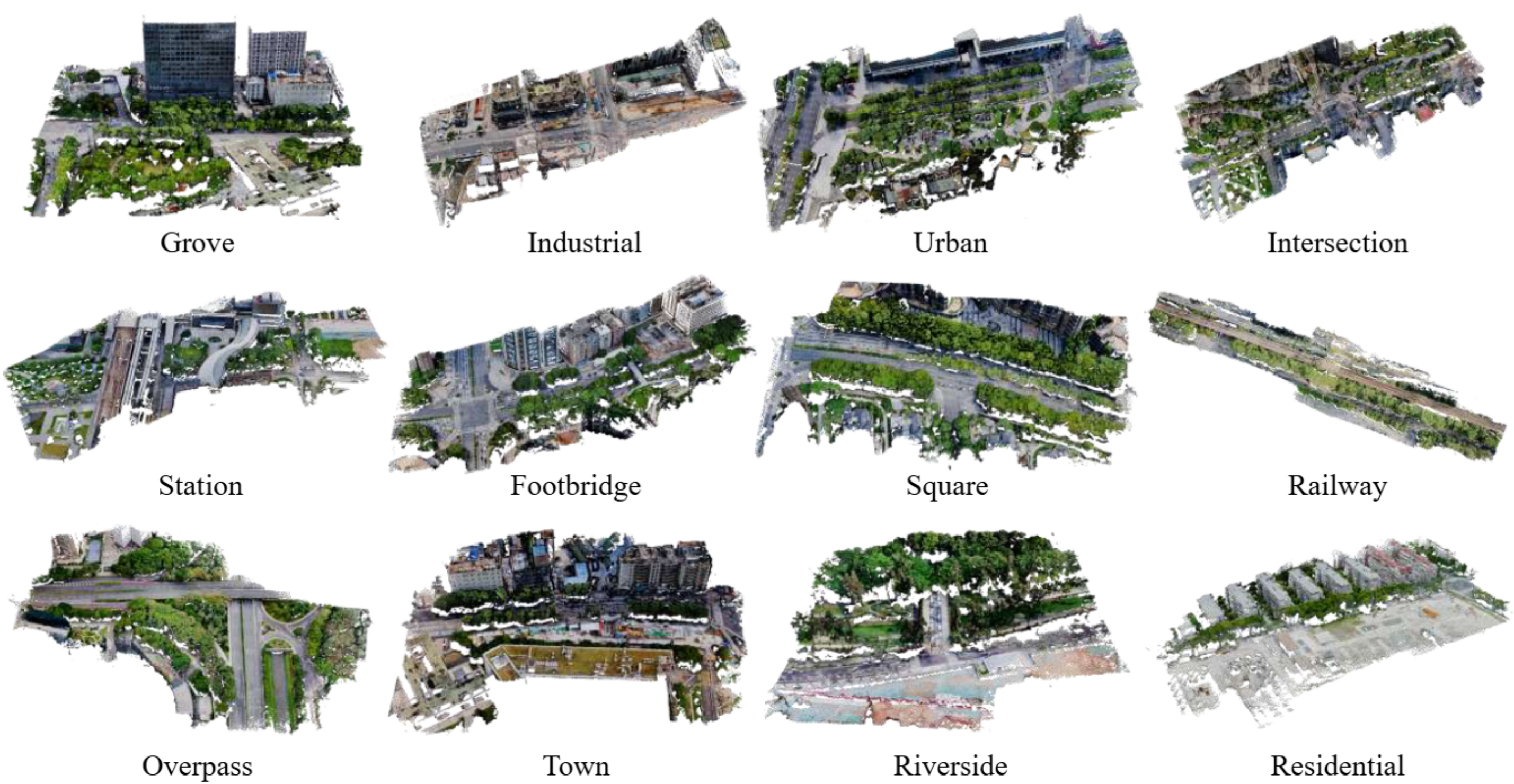} 
\caption{Visualization of representative dense point clouds from 12 selected scenes in our benchmark.}
\label{fig:pointcloud}
\end{figure*}

\begin{table}[t]
\centering
\caption{Data acquisition and reconstruction configuration.}
\label{tab:data_collection}
\small
\begin{tabular}{@{}lp{5cm}@{}}
\toprule
\textbf{Component} & \textbf{Specification} \\
\midrule
UAV platform & DJI flight system with RTK positioning \\
Camera type & Third-party global-shutter RGB camera \\
Resolution & 1280$\times$960 pixels \\
Frame rate & 30\,fps \\
Exposure & Automatic \\
\bottomrule
\end{tabular}
\end{table}

\begin{table*}[t]
\centering
\caption{Quantitative results on Scenes 1--6. Evaluated using PSNR $\uparrow$, SSIM $\uparrow$, and LPIPS $\downarrow$.}
\label{tab:results_s1_s6}
\resizebox{\textwidth}{!}{
\begin{tabular}{l l *{6}{ccc}}
\toprule
Time & Method
& \multicolumn{3}{c}{Town}
& \multicolumn{3}{c}{Residential}
& \multicolumn{3}{c}{Riverside}
& \multicolumn{3}{c}{Grove}
& \multicolumn{3}{c}{Railway}
& \multicolumn{3}{c}{Square}
\\
\cmidrule(lr){3-5}
\cmidrule(lr){6-8}
\cmidrule(lr){9-11}
\cmidrule(lr){12-14}
\cmidrule(lr){15-17}
\cmidrule(lr){18-20}
& 
& PSNR & SSIM & LPIPS
& PSNR & SSIM & LPIPS
& PSNR & SSIM & LPIPS
& PSNR & SSIM & LPIPS
& PSNR & SSIM & LPIPS
& PSNR & SSIM & LPIPS
\\
\midrule

353 hr & NeRF-W  & 19.63 & 0.653 & 0.410  & 20.74 & 0.733 & 0.414  & 17.34 & 0.612 & 0.448  & 17.83 & 0.595 & 0.430  & 16.37 & 0.480 & 0.551  & 19.02 & 0.546 & 0.443 \\
179 hr    & NeRF-OSR  & 18.95 & 0.506 & 0.489  & 20.77 & 0.663 & 0.505  & 18.73 & 0.536 & 0.525  & 17.86 & 0.518 & 0.478  & 15.73 & 0.391 & 0.753  & 18.18 & 0.453 & 0.573 \\
42 hr & GS-W  & 22.27 & 0.787 & 0.161  & \textbf{25.66} & \textbf{0.851} & \textbf{0.144}  & 22.08 & 0.775 & 0.182  & 21.81 & \textbf{0.779} & \textbf{0.144}  & 16.35 & 0.570 & 0.361  & 23.16 & 0.770 & 0.124 \\
46 hr & WildGaussians & \textbf{23.95} & 0.792 & 0.175  & 23.62 & 0.816 & 0.234  & 21.90 & 0.747 & 0.226  & \textbf{22.49} & 0.778 & 0.164  & \textbf{18.34} & 0.606 & 0.382  & 23.62 & 0.746 & 0.168 \\
36 hr & LumiGauss  & 23.59 & \textbf{0.841} & \textbf{0.128}  & 25.14 & \textbf{0.851} & 0.161  & \textbf{23.02} & \textbf{0.809} & \textbf{0.156}  & 21.27 & 0.765 & 0.164  & 17.97 & \textbf{0.624} & \textbf{0.299}  & \textbf{24.31} & \textbf{0.791} & \textbf{0.123} \\

\bottomrule

\end{tabular}
}
\end{table*}

\begin{table*}[t]
\centering
\caption{Quantitative results on Scenes 7--12. Evaluated using PSNR $\uparrow$, SSIM $\uparrow$, and LPIPS $\downarrow$.}
\label{tab:results_s7_s12}
\resizebox{\textwidth}{!}{
\begin{tabular}{l l *{6}{ccc}}
\toprule
Time & Method
& \multicolumn{3}{c}{Footbridge}
& \multicolumn{3}{c}{Industrial}
& \multicolumn{3}{c}{Intersection}
& \multicolumn{3}{c}{Urban}
& \multicolumn{3}{c}{Station}
& \multicolumn{3}{c}{Overpass}
\\
\cmidrule(lr){3-5}
\cmidrule(lr){6-8}
\cmidrule(lr){9-11}
\cmidrule(lr){12-14}
\cmidrule(lr){15-17}
\cmidrule(lr){18-20}
& 
& PSNR & SSIM & LPIPS
& PSNR & SSIM & LPIPS
& PSNR & SSIM & LPIPS
& PSNR & SSIM & LPIPS
& PSNR & SSIM & LPIPS
& PSNR & SSIM & LPIPS
\\
\midrule

356 hr & NeRF-W  & 17.25 & 0.515 & 0.487  & 17.78 & 0.612 & 0.461  & 16.39 & 0.484 & 0.497  & 17.57 & 0.502 & 0.469  & 18.52 & 0.565 & 0.474  & 18.52 & 0.634 & 0.409 \\
181 hr    & NeRF-OSR  & 17.13 & 0.428 & 0.591  & 17.50 & 0.518 & 0.544  & 16.81 & 0.390 & 0.583  & 17.03 & 0.392 & 0.602  & 18.46 & 0.495 & 0.586  & 20.45 & 0.590 & 0.430 \\
43 hr & GS-W  & 17.85 & 0.608 & 0.294  & 19.33 & 0.762 & 0.183  & 19.98 & 0.724 & 0.175  & 19.49 & 0.688 & 0.205  & 22.83 & 0.785 & 0.131  & 22.40 & 0.787 & \textbf{0.127} \\
47 hr & WildGaussians & 17.45 & 0.562 & 0.458  & 16.63 & 0.676 & 0.337  & 21.38 & 0.708 & 0.226  & \textbf{21.50} & 0.705 & 0.218  & 22.23 & 0.756 & 0.177  & \textbf{23.08} & 0.803 & 0.143 \\
35 hr & LumiGauss  & \textbf{20.89} & \textbf{0.742} & \textbf{0.170}  & \textbf{20.34} & \textbf{0.791} & \textbf{0.158}  & \textbf{23.04} & \textbf{0.805} & \textbf{0.127}  & 21.26 & \textbf{0.743} & \textbf{0.164}  & \textbf{23.11} & \textbf{0.805} & \textbf{0.123}  & 22.33 & \textbf{0.813} & \textbf{0.127} \\

\bottomrule
\end{tabular}
}
\end{table*}

\section{Dataset Design Principles}
\label{sec:key_decisions}

UAVLight is designed to isolate light as the primary varying factor while maintaining consistent geometry and views. This is achieved through three key principles, enabling fair benchmarking of reconstruction and relighting methods.

\textbf{Low-altitude Imaging.}  
Flights are conducted at low altitudes with nadir views, where direct sunlight dominates and diffuse components are negligible, yielding physically interpretable lighting changes governed by solar position.

\textbf{Repeated Trajectories.}  
Each scene is captured along identical waypointed paths at several fixed times of day, ensuring comparable viewpoint coverage and parallax across flights.
Crucially, in large-scale outdoor scenes, illumination within a single time slot can be treated as uniform—sun position, cast-shadow directions, and ambient contributions remain stable over the short flight interval.

\textbf{Real-Time Kinematic (RTK)-Based Registration.}  
All cameras use RTK positioning to provide metric-scale priors for SfM and MVS, reducing drift and aligning reconstructions across time in a shared world frame.

\section{Data Collection}
\label{sec:data_collection}

UAVLight is captured through a standardized four-stage pipeline: data acquisition, frame sampling and reconstruction, post-processing, and sunlight estimation. 

\paragraph{Data Acquisition.}
As shown in \Cref{tab:data_collection}, we use a DJI platform with RTK positioning and a global-shutter RGB camera. 
Each scene is captured by repeating the same waypointed trajectory at multiple time slots, ensuring consistent viewpoints while allowing natural variations in sunlight. 
RTK logs provide timestamp, latitude, longitude, and altitude per frame, yielding centimeter-level pose accuracy. 
The camera records 1280$\times$960 images at 30\,fps with automatic exposure, avoiding rolling-shutter artifacts.

\begin{table}[t]
\centering
\footnotesize    
\caption{\textbf{UAVLight scene statistics.} Each scene is captured under multiple natural illumination conditions along repeated flight trajectories.}
\label{tab:dataset_summary}
\begin{tabular}{p{1.4cm} c c c c} 
\toprule
\textbf{Scene} & \textbf{Area (m$^2$)} & \textbf{Traj. (m)} & \textbf{\#Images} & \textbf{\#Illum.} \\
\midrule
Residential        & 37044 & 300 & 267 & 3 \\
Town  & 21624 & 200 & 194 & 8 \\
Grove    & 20900 & 190 & 253 & 11 \\
Railway       & 27280 & 440 & 222 & 5 \\
Riverside   & 22733 & 180 & 224 & 7 \\
Square        & 26985 & 260 & 233 & 6 \\
Footbridge       & 39776 & 350 & 326 & 6 \\
Industrial         & 42130 & 380 & 234 & 3 \\
Intersection         & 31110 & 300 & 278 & 5 \\
Urban         & 32239 & 310 & 239 & 5 \\
Station         & 37600 & 380 & 264 & 6 \\
Overpass         & 39308 & 310 & 290 & 5 \\
Road         & 32120 &  290 & 126 & 3 \\
Park         & 49920 & 520 & 336 & 6 \\
SportsField         & 69204 & 470 & 286 & 3 \\
Industrial2         & 18542 & 260 & 166 & 6 \\
Grove2         & 21505 & 290 & 208 & 10 \\
Road2         & 26790 & 290 & 239 & 5 \\
\bottomrule
\end{tabular}
\label{tab:scenes}
\end{table}

\paragraph{Frame Sampling and Reconstruction.}
To balance coverage and efficiency, we uniformly sample frames at 1\,fps and associate each image with its RTK-GNSS position. 
Building on SfM~\cite{schoenberger2016sfm}, we incorporate geospatial priors into reconstruction via grouped bundle adjustment with RTK constraints~\cite{tagle2024improving}. 
The total energy is:
\begin{equation}
E_{\text{total}} = E_\text{group} + \sum_{i} \kappa_i \left\| \mathbf{c}_i - \mathbf{t}_{\text{RTK}_i} \right\|_2^2,
\end{equation}
where
\begin{equation}
E_\text{group} = \sum_j \rho_j \left( \left\| \pi_g (\mathbf{G}_r, \mathbf{P}_c, \mathbf{X}_k) - \mathbf{x}_{jk} \right\|_2^2 \right).
\end{equation}
Here, \( E_\text{group} \) denotes grouped reprojection error, while the second term softly constrains camera centers to RTK measurements. 
This improves pose accuracy and scale consistency, leading to robust MVS reconstruction and consistent geometry across flights. 
Dense point clouds are also produced (\Cref{fig:pointcloud}).

\begin{figure*}[t]
\centering
\includegraphics[width=0.98\textwidth]{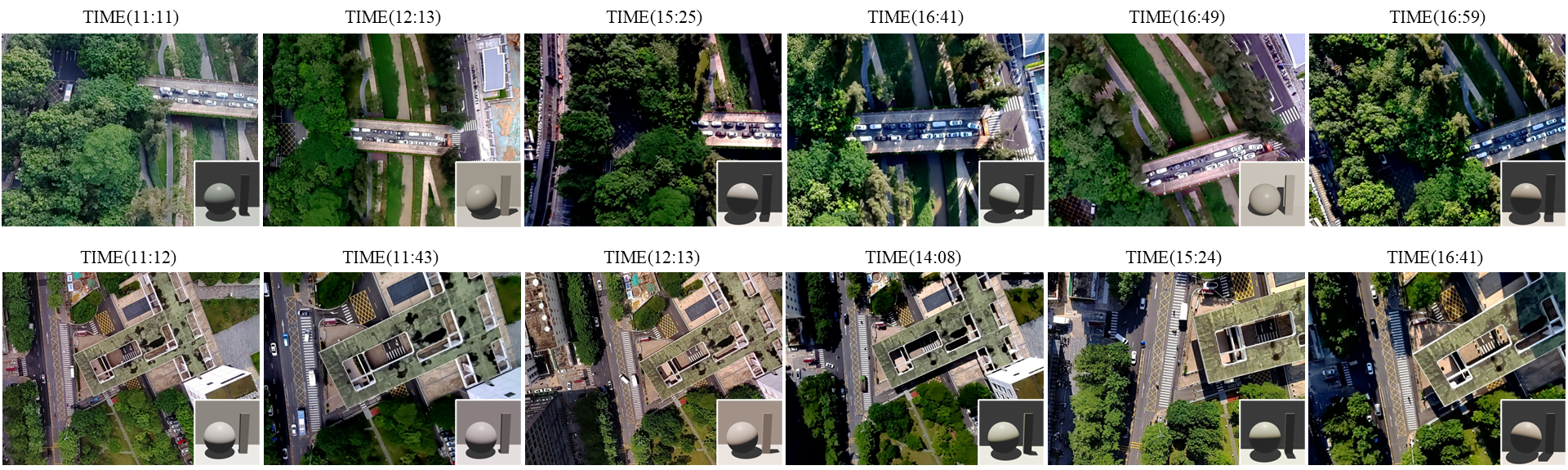}
\caption{Illumination variations across similar viewpoints at different times of day. Bottom-right shows ground-truth sunlight directions from GPS and timestamps.}
\vspace{-0.2cm}
\label{fig:scene_time}
\end{figure*}

\begin{figure}[t]
\centering
\includegraphics[width=1.0\linewidth]{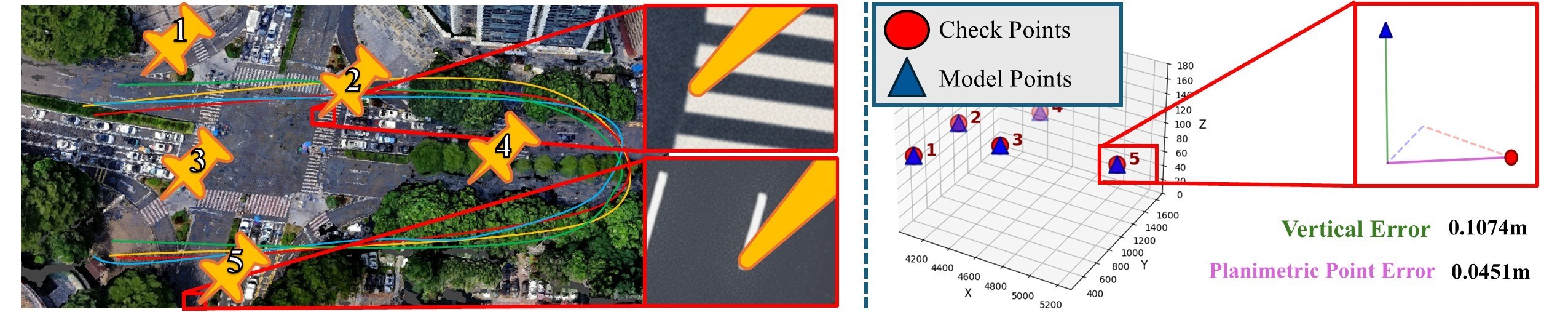} 
\caption{Checkpoint-based geometric validation of UAVLight.}
\label{fig:checkpoint_validation}
\end{figure}

\noindent\textbf{Post Processing.}
We perform manual filtering to remove frames affected by motion blur, extreme exposure, or poor texture (e.g., water surfaces)~\cite{shan2008image, micusik2006radiometric, seitz2006comparison}. 
The remaining images are undistorted using a standard SfM pipeline~\cite{schoenberger2016sfm}, ensuring geometric and photometric consistency for downstream evaluation.

\noindent\textbf{Sunlight Ground Truth.}
UAVLight provides per-time-slot sunlight directions (\Cref{fig:scene_time}). 
Assuming a global directional light source~\cite{holdgeoffroy2019deep}, the sun direction is computed from timestamps and GPS coordinates using the standard solar position algorithm~\cite{reda2008solar}. 
Given time \( t \), longitude \( \lambda \), and latitude \( \phi \), we compute solar altitude \( \alpha_\text{sun} \) and azimuth \( \gamma_\text{sun} \), with zenith angle \( \theta_\text{sun} = 90^\circ - \alpha_\text{sun} \). 
The unit sun direction in the local ENU frame is:
\begin{equation}
\label{eq:sun_vector_enu}
\begin{aligned}
s_{\text{E}} &= \sin(\gamma_\text{sun}), \\
s_{\text{N}} &= \cos(\gamma_\text{sun}), \\
s_{\text{U}} &= \sin(\alpha_\text{sun}) = \cos(\theta_\text{sun}).
\end{aligned}
\end{equation}

This direction is transformed to the COLMAP frame via:
\begin{equation}
\label{eq:transform_to_colmap}
\mathbf{s}_{\text{Colmap}} = \mathbf{R} \, \mathbf{s}_{\text{ENU}}.
\end{equation}
Such physically grounded sunlight annotations provide reliable supervision for lighting estimation~\cite{holdgeoffroy2019deep,zhang2023relitnerf, GaSLight}, inverse rendering~\cite{du2024gs, GI-GS}, and relightable reconstruction~\cite{lumigauss}.

\paragraph{Geometry Check.}
We validate geometry using checkpoint-based survey measurements (\Cref{fig:checkpoint_validation}). 
The reference is anchored by RTK camera priors and ground control points, providing independent metric constraints. 
As shown in \Cref{tab:geometry_summary}, the average vertical and planimetric errors are 10.31\,cm and 11.83\,cm, respectively, consistent with standard UAV photogrammetry.

\begin{figure}[!t]
\centering
\includegraphics[width=0.98\linewidth]{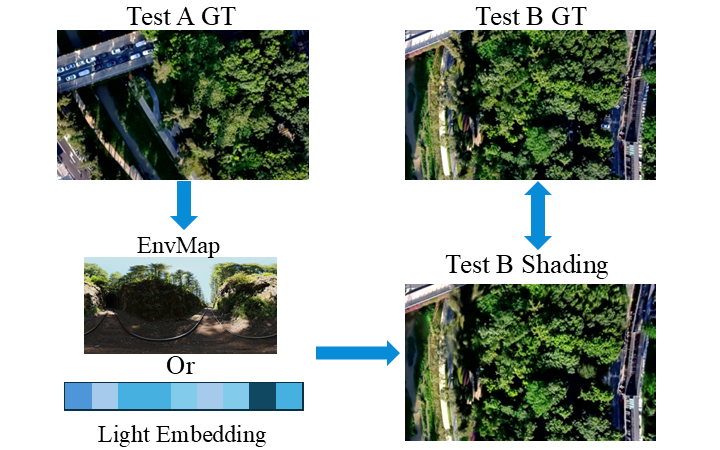}
\vspace{-0.2cm}
\caption{Paired cross-light protocol. Lighting is estimated from one view and applied to another captured at the same time slot.}
\label{fig:experiment}
\end{figure}

\section{Scenes}
\label{sec:scenes}
UAVLight comprises 18 real-world outdoor scenes covering large spatial areas with diverse geometry, materials, and environments. 
This diversity leads to rich illumination effects, including cast shadows, specular highlights, diffuse shading, and global illumination. 
The dataset spans the following scene types (\Cref{tab:scenes}):

\begin{itemize}
    \item \textbf{Natural Vegetation}: (\textit{Park}, \textit{Grove}, \textit{Grove2}) Dense foliage introduces heavy occlusion, dappled lighting, soft shadows, and spatially varying reflectance.
    
    \item \textbf{Urban and Residential}: (\textit{Residential}, \textit{Urban}, \textit{Town}) Complex geometry and heterogeneous materials (e.g., facades, rooftops, windows) produce deep cast shadows and structured shadow boundaries.
    
    \item \textbf{Open Public Spaces}: (\textit{Square}, \textit{SportsField}) Large open areas with simple geometry and uniform materials yield clear shadow boundaries and consistent lighting response.
    
    \item \textbf{Industrial Sites}: (\textit{Industrial}, \textit{Industrial2}) Repetitive structures and metallic surfaces introduce anisotropic highlights and strong directional shading.
    
    \item \textbf{Transportation Infrastructure}: (\textit{Road}, \textit{Road2}, \textit{Intersection}, \textit{Footbridge}, \textit{Station}, \textit{Overpass}, \textit{Railway}) Multi-level layouts with wide ground regions exhibit long-range shadows and sharp illumination transitions.
    
    \item \textbf{Riverside}: (\textit{Riverside}) Water reflections, vegetation scattering, and bridge shadows with non-Lambertian effects.
\end{itemize}

\begin{table}[t]
\centering
\caption{Summary of checkpoint-based geometric accuracy (average vertical and planimetric errors).}
\label{tab:geometry_summary}
\resizebox{\columnwidth}{!}{%
\begin{tabular}{cccccc}
\toprule
\#Scenes & Altitude (m) & \#Checkpoints & Oblique : Nadir & Vertical Error (cm) & Planimetric Point Error (cm) \\
\midrule
18 & 80--100 & $\approx$10 / scene & All Nadir & $10.31$ & $11.83$ \\
\bottomrule
\end{tabular}}
\end{table}

\begin{figure*}[!ht]
\centering
\includegraphics[width=0.95\textwidth]{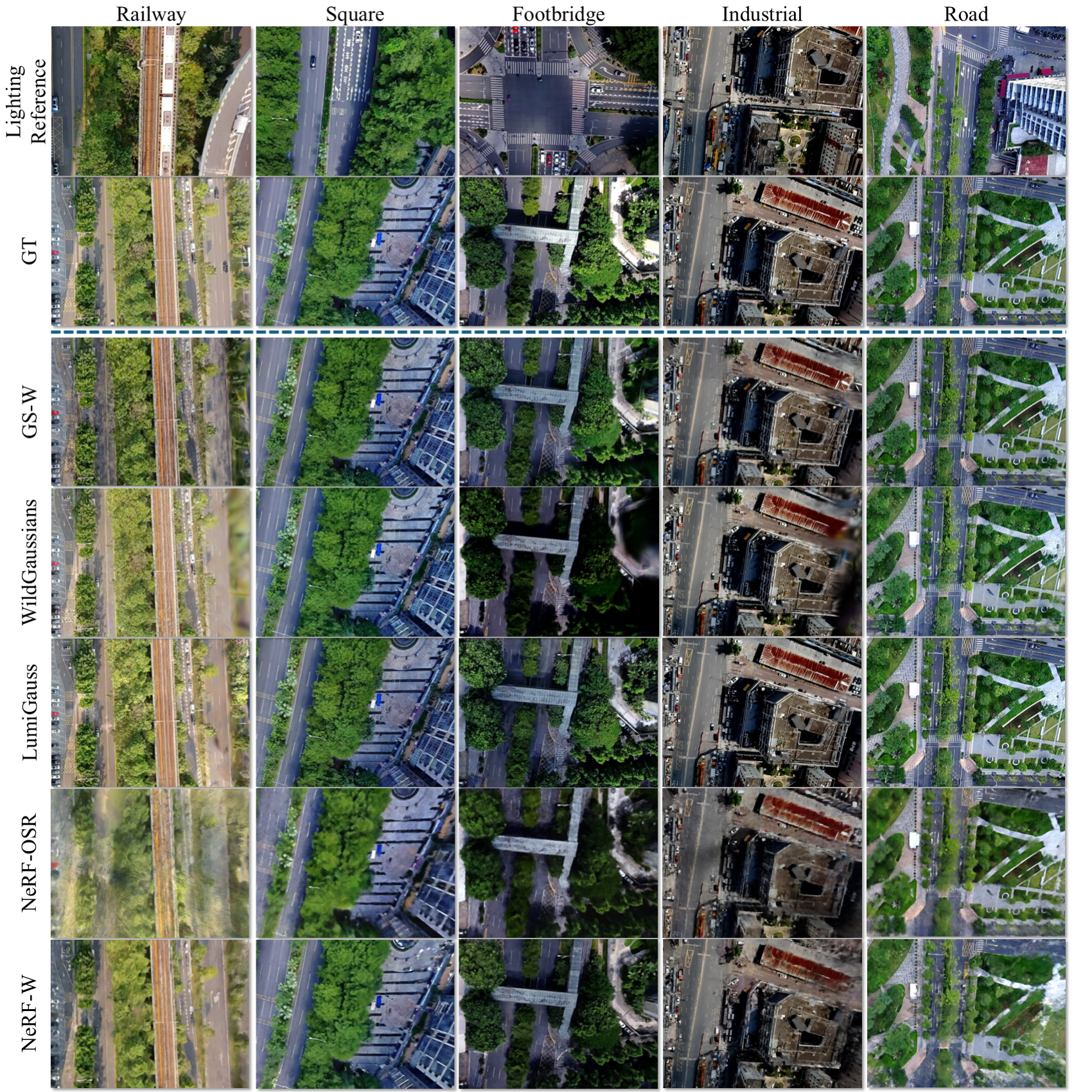} 
\caption{Visualization of the reconstruction results from different baselines on five UAVLight scenes.}
\label{fig:compare}
\vspace{-0.2cm}
\end{figure*}

\begin{figure*}[!ht]
\centering
\includegraphics[width=0.95\textwidth]{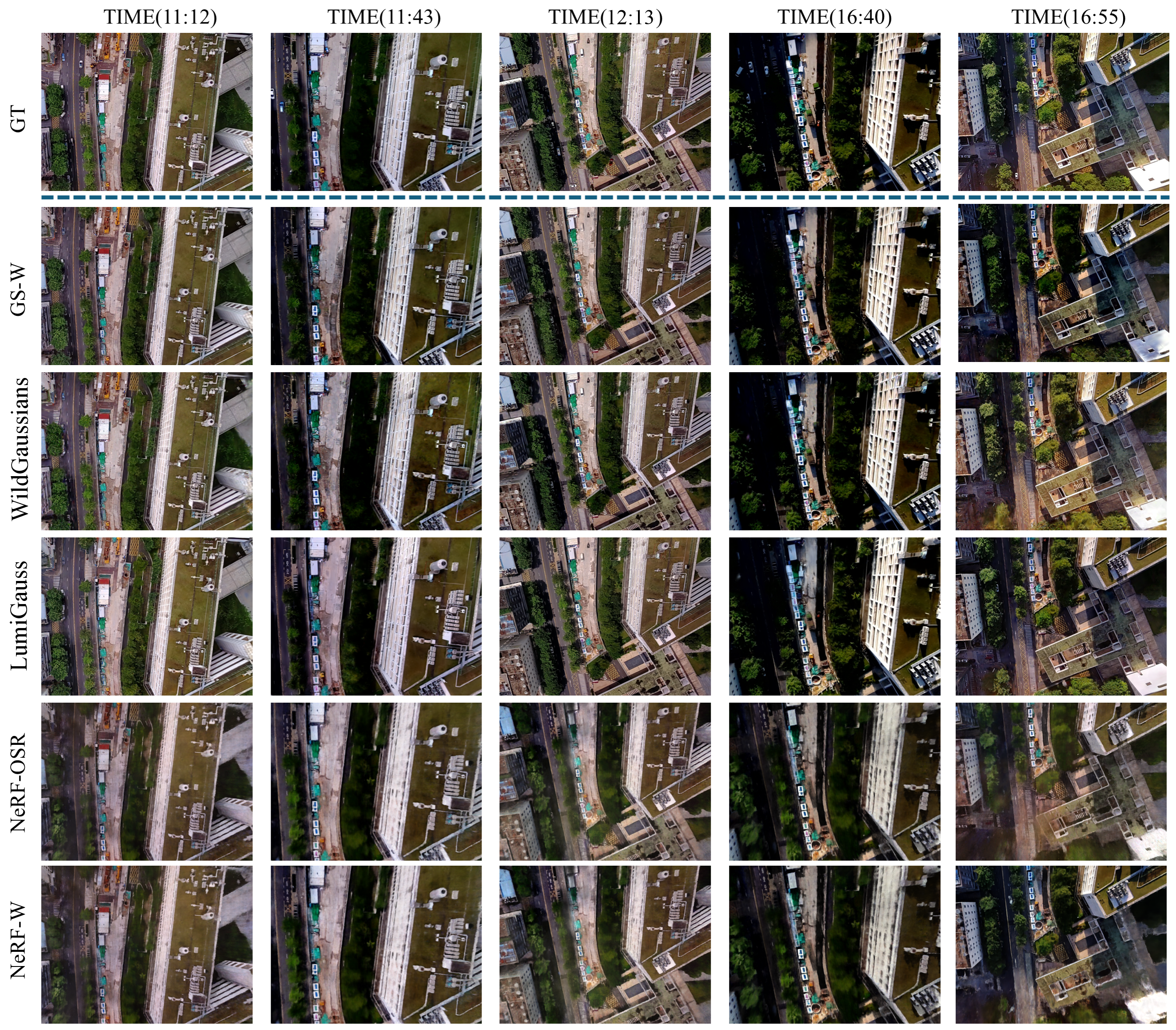} 
\caption{Visualization of the reconstruction results from different baselines on Town with five different time slots.}
\label{fig:compare2}
\end{figure*}


\section{Experiments}
\label{sec:experiments}

We present a benchmark for inverse rendering that enables fair evaluation of 3D reconstruction under various lighting conditions without requiring external environment maps.

\subsection{Experiment Design}
\label{sec:app_experiments}

Existing evaluation protocols generally fall into two categories.
(1) \cite{MartinBrualla2021NeRFW}, and its followers adopt a half-split protocol: estimate lighting from one half of the views and evaluate on the other half. However, using only half of the views may provide incomplete lighting cues, causing the learned embedding to overfit view-specific appearance.
(2) Methods such as \cite{Chen2022NeRFOSR} evaluated with calibrated environment maps, which are physically grounded but are expensive to capture for all time slots and difficult to apply to large outdoor areas.
As shown in~\Cref{fig:experiment}, to support both implicit and explicit-illumination approaches, we adopt a paired cross-illumination setup in which lighting is estimated from one subset of views and evaluated on another within the same time slot. This ensures consistent geometry while isolating illumination as the only varying factor without GT envmap.

\subsection{Data Splits and Reproducibility}
To ensure fair comparison and reproducibility, UAVLight follows a standardized train/val/test split across different time slots, avoiding temporal leakage. 
For each test slot, camera views are further divided into two subsets, \(A_t\) and \(B_t\), matched by altitude and view angle to enable unbiased cross-illumination evaluation. We release all configuration details—fixed random seeds, A/B view indices, exposure normalization parameters, and an official evaluation script—to guarantee consistent and repeatable results.

\subsection{Baselines}
We evaluate five representative baselines, which fall into two broad categories:
\textbf{Implicit illumination modeling.}
These approaches encode lighting variations implicitly through learned latent features, without explicitly recovering scene illumination. This category includes GS-W \cite{Zhang2024GSW}, WildGaussians \cite{Kulhanek2024WildGaussians}, and NeRF-W \cite{MartinBrualla2021NeRFW}.
\textbf{Explicit illumination modeling.}
These inverse-rendering-based methods recover shared geometry and materials across time, while estimating per-capture lighting for relighting. This category includes LumiGauss \cite{lumigauss} and NeRF-OSR \cite{Chen2022NeRFOSR}.

\section{Results}
We evaluate all baselines across 12 representative scenes in UAVLight.
Our goal is to examine how existing 3D reconstruction and inverse rendering approaches behave under real-world illumination variations, which previous benchmarks have struggled with. Quantitative results are summarized in \Cref{tab:results_s1_s6,tab:results_s7_s12}, and visual comparisons are presented in \Cref{fig:compare,fig:compare2}. The remaining 6 scenes and additional experimental analysis are detailed in the supplement.

\subsection{Quantitative Analysis}
Across all scenes, clear and interpretable trends emerge.
Gaussian-based methods achieve the strongest performance in standard metrics (PSNR \cite{hore2010image}, SSIM \cite{wang2004image}, LPIPS \cite{zhang2018unreasonable}), confirming their stability under multi-view reconstruction.
More importantly, explicit illumination models (LumiGauss \cite{lumigauss}) consistently outperform implicit ones (GS-W \cite{Zhang2024GSW}, WildGaussians \cite{Kulhanek2024WildGaussians}, NeRF-W \cite{MartinBrualla2021NeRFW}) when evaluated across different illumination slots. This gap highlights a key difficulty revealed by UAVLight: methods that entangle lighting with color struggle to maintain geometry-material consistency under illumination changes, whereas explicit decompositions provide more reliable supervision.

\subsection{Qualitative Analysis}
The qualitative results further illustrate the role of UAVLight in exposing illumination-dependent failure modes.
\emph{Footbridge} in \Cref{fig:compare} shows that implicit methods, though sometimes sharper around high-frequency areas, such as shadow boundaries, often imprint shadows into albedo or distort geometry when light changes across time of day.
In contrast, explicit illumination methods produce more coherent soft shadows.
\Cref{fig:compare2} shows that these trends persist across diverse lighting conditions. In particular, at TIME (16:55), GS-W produces sharper shadows, whereas LumiGauss yields shading closer to ground truth.

\subsection{Overall Observation}
The results show that UAVLight not only distinguishes the strengths of Gaussian-based reconstruction, but also reveals the fundamental trade-offs between implicit and explicit illumination modeling.  
Explicit approaches exhibit stronger disentanglement and more stable multi-light reconstruction, while implicit ones capture certain high-frequency effects but are more prone to entanglement-induced artifacts. More results are provided in the supplemental materials.

\section{Conclusion}
In summary, UAVLight establishes the first benchmark for multi-lighting reconstruction in real-world UAV scenes. Through capturing consistent frames across natural lighting variations, it enables a fair comparison of implicit and explicit lighting models.
Future extensions of UAVLight include expanding illumination diversity (clouds, seasons, atmospheric variation), and developing real-to-sim paired versions for outdoor simulation and embodied-AI research.

{
    \small
    \bibliographystyle{ieeenat_fullname}
    \bibliography{main}
}

\clearpage
\setcounter{page}{1}
\maketitlesupplementary

\section*{Overview}

This supplementary material provides additional qualitative and quantitative results complementing the UAVLight benchmark paper. It is organized into four parts, and we also provide a video file containing extended visualizations across time slots and viewpoints.

\begin{figure*}[!ht]
\centering
\includegraphics[width=\linewidth]{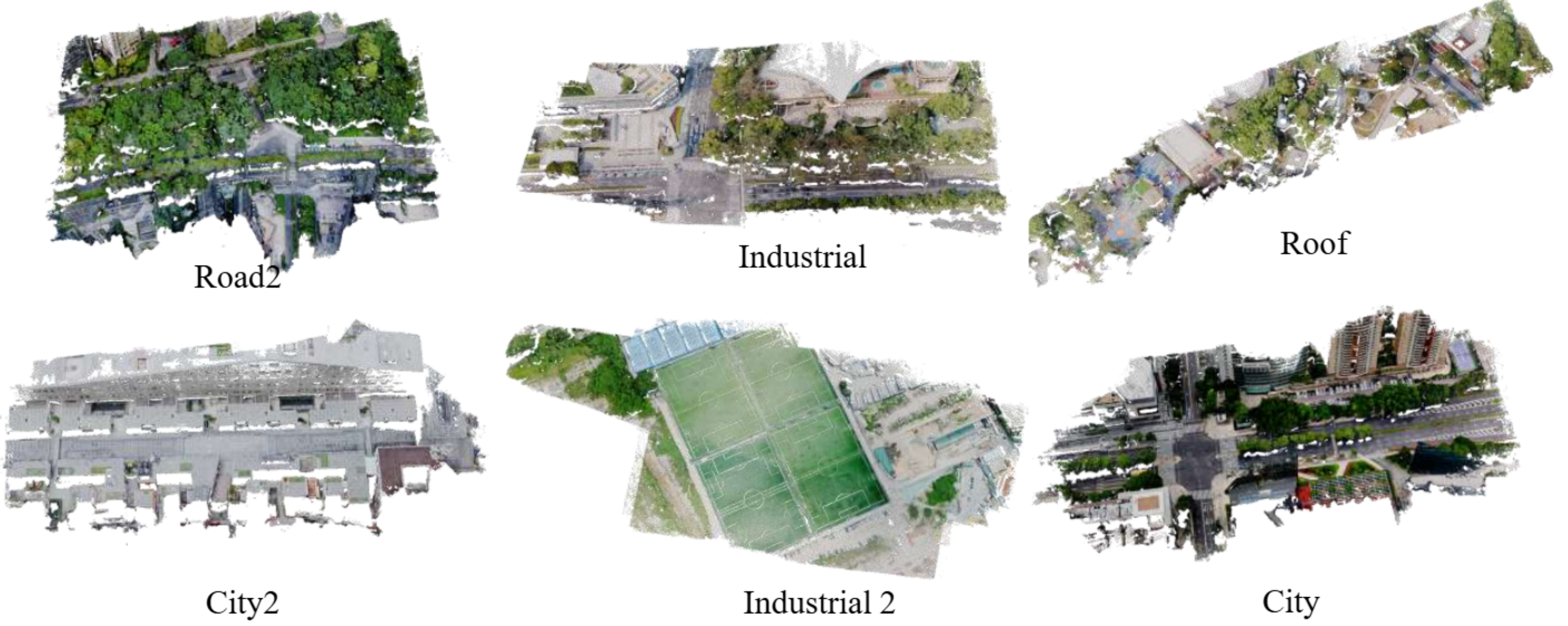}
\caption{
\textbf{Dense point cloud visualizations for the remaining six scenes in UAVLight.}
These reconstructions exhibit diverse structural layouts, including vegetation, metallic structures, and transportation infrastructure, serving as ground-truth geometric references under changing natural illumination.
}
\label{fig:supp_pointcloud}
\vspace{-0.2cm}
\end{figure*}

\begin{figure*}[!ht]
\centering
\includegraphics[width=\linewidth]{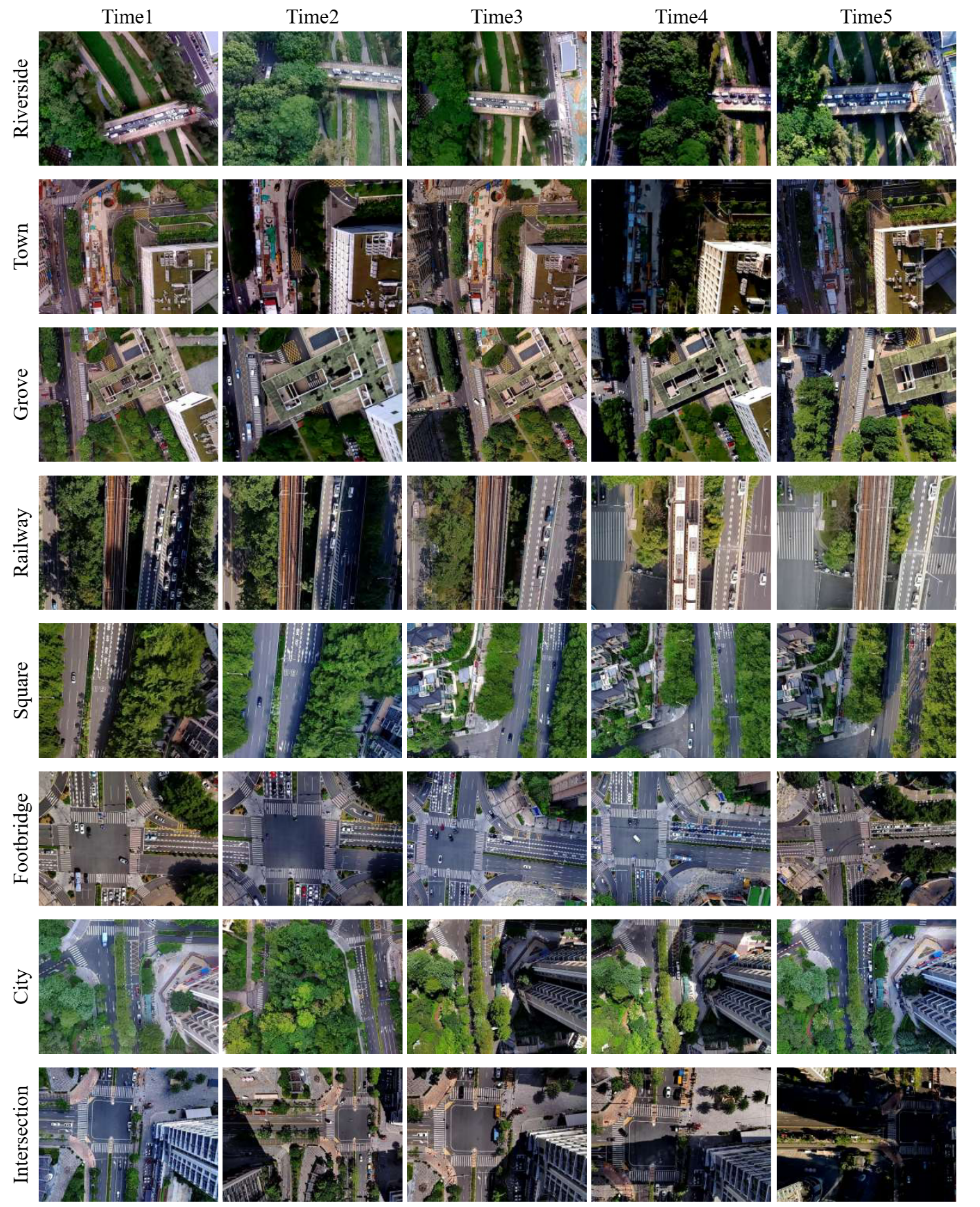}
\caption{
\textbf{Representative multi-time-slot visualizations (Part I).}
Each scene is shown under 3–5 time slots captured along identical waypointed UAV trajectories.  
Shadow displacement, shading gradients, and ambient-light variations illustrate the natural illumination diversity captured by UAVLight.
}
\label{fig:supp_compare_1}
\vspace{-0.2cm}
\end{figure*}

\begin{figure*}[!ht]
\centering
\includegraphics[width=\linewidth]{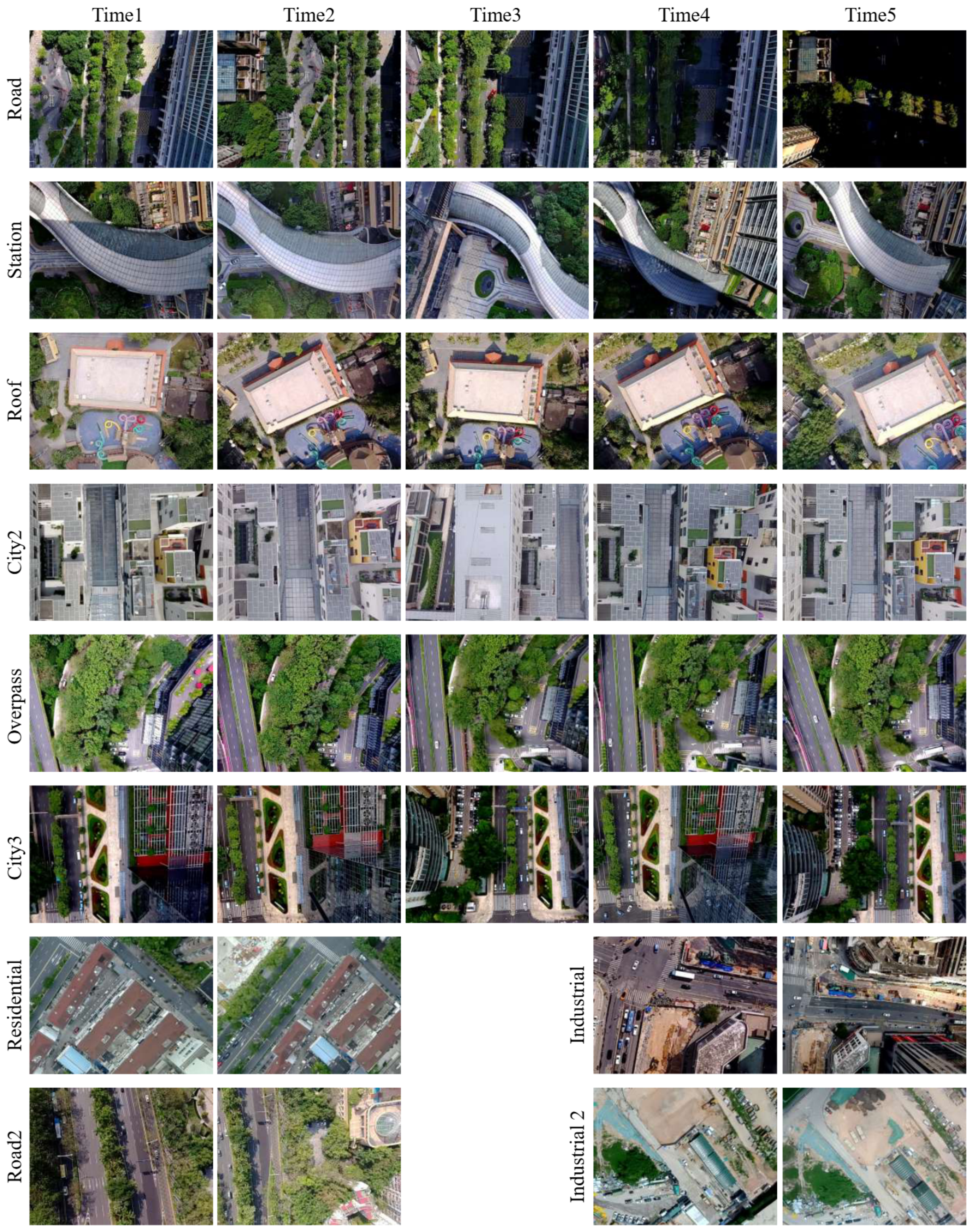}
\caption{
\textbf{Representative multi-time-slot visualizations (Part II).}
Additional scenes showing consistent geometry under varying sunlight conditions.  
These examples highlight the challenges posed by specularities, vegetation, and long cast shadows.
}
\label{fig:supp_compare_2}
\vspace{-0.2cm}
\end{figure*}

\begin{figure*}[!ht]
\centering
\includegraphics[width=\linewidth]{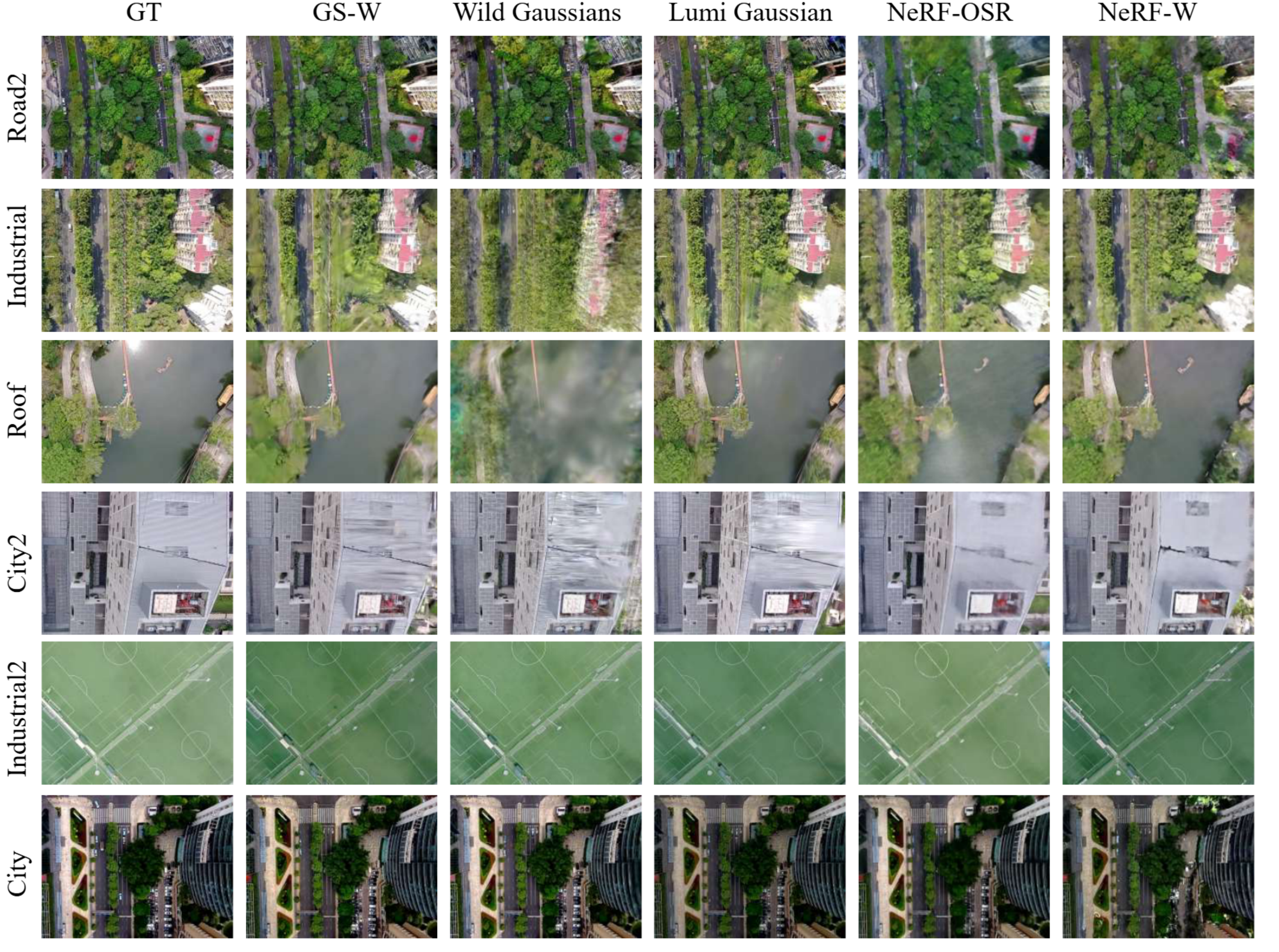}
\caption{
\textbf{Additional comparisons on the remaining six scenes.}
We present the rendering results from all baselines, complementing the main paper and providing full-scene coverage of implicit and explicit illumination models across UAVLight.
}
\label{fig:compare_supp}
\vspace{-0.2cm}
\end{figure*}

\begin{itemize}
    \item \textbf{Part 1: Additional Dense Point Cloud Visualizations.}  
    We provide dense point cloud renderings for the remaining six scenes not shown in the main paper, illustrating large-scale geometry and illumination-sensitive regions (Fig.~\ref{fig:supp_pointcloud}).

    \item \textbf{Part 2: Multi–Time-Slot Visualizations Across All Scenes.}  
    Figs.~\ref{fig:supp_compare_1} and \ref{fig:supp_compare_2} show 3–5 representative time slots per scene, demonstrating how natural sunlight variations affect appearance while geometry and viewpoints remain fixed.

    \item \textbf{Part 3: Additional Baseline Comparisons.}  
    We provide qualitative comparisons of all evaluated baselines on the remaining six scenes, as shown in \Cref{fig:compare_supp}. These examples extend the main paper’s analysis of explicit versus implicit illumination modeling.

    \item \textbf{Part 4: Extended Quantitative Results.}
    We report full PSNR, SSIM, and LPIPS results for the remaining six scenes, complementing Tables~4 and~5 in the main paper.  
    The metrics are presented in two tables, each covering three scenes (Road2, Industrial, Roof in Table~\ref{tab:extended_results_part1}; City2, Industrial2, City in Table~\ref{tab:extended_results_part2}).  
    These results complete the evaluation over all 18 UAVLight scenes and further demonstrate the benchmark’s illumination diversity and consistency across time.
    
\end{itemize}

These additions offer a complete view of UAVLight’s behavior across all 18 scenes and multi-time-of-day captures, further supporting its role as a controlled-yet-natural benchmark for illumination-robust reconstruction.

\section*{Reproducibility Details}

To ensure full reproducibility, we provide the following implementation and configuration details used across all experiments:

\begin{itemize}
    \item \textbf{Hardware.}  
    NVIDIA A800 GPUs, 80GB VRAM for all baseline methods.

    \item \textbf{Software.}  
    PyTorch 2.4.1, CUDA 11.8, COLMAP 3.7.

    \item \textbf{Dataset Release.}  
    We release camera poses, A/B split indices per time slot, undistorted images, sunlight vectors, dense pointcloud, and evaluation scripts.

\end{itemize}

\begin{table*}[t]
\centering
\small
\setlength{\tabcolsep}{6pt}
\begin{tabular}{l|ccc|ccc|ccc}
\toprule
\multirow{2}{*}{\textbf{Method}} 
& \multicolumn{3}{c|}{\textbf{Road2}} 
& \multicolumn{3}{c|}{\textbf{Industrial}} 
& \multicolumn{3}{c}{\textbf{Roof}} \\
& PSNR & SSIM & LPIPS 
& PSNR & SSIM & LPIPS
& PSNR & SSIM & LPIPS \\
\midrule

\textbf{LumiGauss} 
& 22.24 & 0.750 & 0.163
& 17.42 & 0.676 & 0.223
& 19.79 & 0.654 & 0.288 \\

\textbf{WildGaussians} 
& 21.60 & 0.692 & 0.223
& 17.49 & 0.580 & 0.338
& 16.91 & 0.532 & 0.555 \\

\textbf{GaussianWild} 
& 21.34 & 0.721 & 0.170
& 18.13 & 0.622 & 0.285
& 19.29 & 0.587 & 0.394 \\

\textbf{NeRF-OSR} 
& 16.53 & 0.369 & 0.581
& 18.85 & 0.513 & 0.555
& 18.17 & 0.462 & 0.691 \\

\textbf{NeRF-W} 
& 16.93 & 0.431 & 0.493
& 16.45 & 0.534 & 0.481
& 18.18 & 0.516 & 0.504 \\

\bottomrule
\end{tabular}
\caption{\textbf{Extended quantitative results (Part 1)} on Road2, Industrial, and Roof.  
Metrics are reported separately for PSNR, SSIM, and LPIPS (no slash).}
\label{tab:extended_results_part1}
\end{table*}

\begin{table*}[t]
\centering
\small
\setlength{\tabcolsep}{6pt}
\begin{tabular}{l|ccc|ccc|ccc}
\toprule
\multirow{2}{*}{\textbf{Method}} 
& \multicolumn{3}{c|}{\textbf{City2}} 
& \multicolumn{3}{c|}{\textbf{Industrial2}} 
& \multicolumn{3}{c}{\textbf{City}} \\
& PSNR & SSIM & LPIPS 
& PSNR & SSIM & LPIPS
& PSNR & SSIM & LPIPS \\
\midrule

\textbf{LumiGauss} 
& 22.69 & 0.820 & 0.148
& 20.03 & 0.850 & 0.188
& 19.77 & 0.780 & 0.155 \\

\textbf{WildGaussians} 
& 21.21 & 0.759 & 0.215
& 24.80 & 0.909 & 0.121
& 20.82 & 0.749 & 0.180 \\

\textbf{GaussianWild} 
& 21.30 & 0.771 & 0.167
& 22.33 & 0.892 & 0.115
& 20.72 & 0.752 & 0.163 \\

\textbf{NeRF-OSR} 
& 19.38 & 0.552 & 0.535
& 21.47 & 0.789 & 0.378
& 19.71 & 0.595 & 0.413 \\

\textbf{NeRF-W} 
& 19.62 & 0.602 & 0.460
& 20.49 & 0.825 & 0.339
& 17.42 & 0.619 & 0.409 \\

\bottomrule
\end{tabular}
\caption{\textbf{Extended quantitative results (Part 2)} on City2, Industrial2, and City.  
Metrics are reported separately for PSNR, SSIM, and LPIPS.}
\label{tab:extended_results_part2}
\end{table*}

\end{document}